\documentclass[sn-mathphys, iicol]{sn-jnl}

\usepackage[utf8]{inputenc}
\usepackage{xcolor}
\usepackage{amssymb}
\usepackage{float}
\usepackage{amsmath}
\usepackage{hyperref}
\usepackage{graphicx}
\usepackage{caption}
\usepackage{subcaption}
\usepackage{booktabs} 
\usepackage{algorithm,algpseudocode}
\usepackage{array}
\usepackage{subcaption}

\normalbaroutside
\usepackage{todonotes}

\usepackage{cleveref}

\begin{document}

\title[]{Wind speed super-resolution and validation: from ERA5 to CERRA via diffusion models}

\author*[1]{\fnm{Fabio} \sur{Merizzi}}\email{fabio.merizzi@unibo.it}

\author[1]{\fnm{Andrea} \sur{Asperti}}\email{andrea.asperti@unibo.it}

\author[1]{\fnm{Stefano} \sur{Colamonaco}}\email{stefano.colamonaco@studio.unibo.it}


\affil[1]{\orgdiv{Department of Informatics: Science and Engineering (DISI)}, \orgname{University of Bologna}, \orgaddress{\street{Mura Anteo Zamboni 7}, \city{Bologna}, \postcode{40126},
\country{Italy}
}}



\abstract{
The Copernicus Regional Reanalysis for Europe, CERRA, is a high-resolution regional reanalysis dataset for the European domain. In recent years it has shown significant utility across various climate-related tasks, ranging from forecasting and climate change research to renewable energy prediction, resource management, air quality risk assessment, and the forecasting of rare events, among others. Unfortunately, the availability of CERRA is lagging two years behind the current date, due to constraints in acquiring the requisite external data and the intensive computational demands inherent in its generation. As a solution, this paper introduces a novel method using diffusion models to approximate CERRA downscaling in a data-driven manner, without additional informations. By leveraging the lower resolution ERA5 dataset, which provides boundary conditions for CERRA, we approach this as a super-resolution task. Focusing on wind speed around Italy, our model, trained on existing CERRA data, shows promising results, closely mirroring original CERRA data. Validation with in-situ observations further confirms the model's accuracy in approximating ground measurements. }

\maketitle
\markboth{}{}

\noindent \textbf{keywords:} super-resolution, statistical downscaling, diffusion models, weather reanalysis

\section{Introduction}

In this study, we propose the utilization of super-resolution neural network architectures for the downscaling of reanalysis datasets in the field of meteorology and climate science. Recent advancements in super-resolution techniques have significantly enhanced their performance across a variety of applications, rivalling traditional physics-based models. This promising development suggests that these neural architectures could offer a more effective and practical approach in the field of climate data analysis, potentially transforming our ability to obtain high-resolution insights from coarse-scale meteorological data.

\begin{figure}[h]
    \centering
    \begin{subfigure}[b]{0.49\linewidth}
        \includegraphics[width=\linewidth]{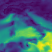}
        \caption{ERA5}
    \end{subfigure}
    \hfill 
    \begin{subfigure}[b]{0.49\linewidth}
        \includegraphics[width=\linewidth]{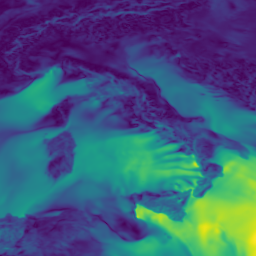}
        \caption{CERRA}
    \end{subfigure}
    \caption{Comparison of wind speed over Italy between ERA5 and CERRA}
    \label{fig:era-cerra-intro}
\end{figure}

The downscaling models typically used for reanalysis in meteorology share a problem formulation similar to super-resolution, where a low-resolution image is transformed into a high-resolution counterpart. This similarity opens up the possibility of utilizing data-driven neural methods to approximate the performance of physics-inspired downscaling models. However, current downscaling models are often hindered by high computational costs and typically require a vast amount of additional information for effective downscaling. This additional information frequently depends on various sources, which can delay the availability of the data. Coupled with the high computational demands, these factors tend to postpone the availability of downscaling results, often lagging years behind the current date and usually falling behind more available lower-resolution reanalysis. This delay can impair the ability to access fine-resolution datasets of recent events, thus delaying the response of the scientific community. This was particularly evident in the case of the Italian peninsula, especially with respect to the May 2023 Emilia-Romagna floods and the November 2023 Ciarán Storm.

Our methodology involves creating a framework to evaluate the efficacy of state of the art neural super-resolution models in a recognized downscaling task. We selected ERA5 \cite{Hersbach_ERA5_2018} and CERRA \cite{Schimanke_CERRA_2021} as our low-resolution and high-resolution datasets, respectively. This choice is particularly advantageous because CERRA is derived using ERA5 as a baseline, establishing a robust connection between the two datasets and minimizing their differences. For our experimental setup, we focused on a region encompassing the Italian peninsula, bounded by the following geographical coordinates: North at 47.75°, South at 35°, East at 18.75°, and West at 6°. We chose a training period of 10 years, from 2010 to 2019, with a temporal resolution of 3 hours, and selected two years, 2009 and 2020, as our test set. Our primary strategy employs a conditioned diffusion model for downscaling, a model that has recently demonstrated top-tier performance in various applications. In addition to the diffusion model, we evaluated a range of other well-known architectures for comparative analysis. To validate our results, we incorporated in-situ observations from ground stations, which aided in assessing the real-world accuracy and effectiveness of our downscaling approach.

With our research we aim to demonstrate the feasibility of neural super-resolution architectures in the data-driven downscaling of reanalysis datasets. At the time of writing, data availability issues made CERRA reanalysis lag more than 2 years behind ERA5, impairing the research on current events. Our objective is to harness the capabilities of our super-resolution, diffusion-based approach to provide researchers with rapid and effective access to downscaled data, facilitating a swift response to current events and eliminating lengthy delays in data availability. We also seek to demonstrate the feasibility of neural methods for downscaling altogether, with possible future integration with physics inspired model to create more efficient and better all-around performing models. We argue that the application of super-resolution models in downscaling could significantly aid in the understanding and prediction of weather and climate patterns at a local scale, proving particularly beneficial for specialized areas such as agriculture, urban planning, and disaster management.

In this paper, we start by examining the correlation between the downscaling problem and the super-resolution problem. Following this, we provide an overview of the super-resolution research field, with an emphasis on diffusion models. We then detail the datasets utilized in our experiments, specifically ERA5, CERRA, and IGRA V2 \cite{durre2016igra}. Subsequently, we introduce our methodology and delve into the specifics of our diffusion model, including an analysis of training and evaluation processes, as well as comparisons with competing models. Our validation framework is then described. Finally, we outline our experimental procedures and discuss the results obtained, concluding with reflections on potential future work and applications in this field.



\section{Super-Resolution for downscaling}
Super-Resolution (SR) and statistical downscaling, while distinct in their methodologies and objectives, share a common goal of refining data resolution, particularly in the context of weather datasets. Super-resolution, primarily used in image processing, involves increasing the resolution of an image by adding more pixels and thereby enhancing the detail and clarity of the image.
In recent years, super-resolution has played a pivotal role in the evolution of deep learning models, demonstrating significant improvements in performance over traditional non-neural methods.

In contrast, statistical downscaling in meteorology is concerned with translating large-scale climate or weather data (often from global climate models such as ERA5) into more localized, high-resolution data. This is crucial for understanding local climate impacts, with the global models often able to capture small scale phenomenons in detail. Statistical downscaling uses statistical methods to establish relationships between large-scale atmospheric variables and local weather phenomena. 

With both tasks aiming to provide higher resolution outputs, we argue that a deep learning approach at super-resolution can be applied to the downscaling of metereological variables. 

\subsection{Problem Definition}

Super-resolution is a process in digital image manipulation that enhances image resolution. This problem can be defined as an ill posed inverse problem, where the task is to invert an unknown degradation function applied to a high resolution image. Given a low resolution image denoted by $y$ and the corresponding high resolution denoted by $x$, we define the degradation process as:
\begin{equation}
    y = \Theta(x, \rho_n)
    \label{eq:problem_definition}
\end{equation}

where $\Theta$ is the degradation function, and $\rho_n$ denotes the degradation parameters (such as the scaling factor, noise, etc.). In a real-world scenario, only $y$ is available while no information about the degradation process or the degradation parameters $\rho$ exists. Super-resolution seeks to nullify the
degradation effect and recovers an approximation $\hat{x}$ of the ground-truth image $x$ as:
\begin{equation}
    \hat{x} = \Theta^{-1}(y, \rho_r)
    \label{eq:inverse_problem}
\end{equation}
where $\rho_r$ are the parameters for the function $\Theta^{-1}$. The degradation process is unknown and can be quite complex. It can be affected by several factors, such as noise, compression,
blur and other artifacts. Therefore, most research works \cite{10.1145/3390462,LEPCHA2023230,bashir2022comprehensive} prefer the following
degradation model over that of Equation \ref{eq:problem_definition}:
\begin{equation}
    y = (x \otimes k) \downarrow_s + n  
    \label{eq:conv_degradation}
\end{equation}

where $k$ is the blurring kernel and $(x \otimes k)$ is the convolution operation between the HR image and the blur kernel, $\downarrow_s$ is a downsampling operation with a scaling factor $s$. 

Let us consider now the task of statistical downscaling in weather sciences. Statistical downscaling is a technique employed to convert broad-scale meteorological data, typically obtained from global climate models or reanalysis products, into finer, local-scale weather information. This approach relies on establishing statistical relationships between large-scale atmospheric variables (like temperature, pressure, and wind) and local-scale weather phenomena. This process is usually implemented via a sophisticated physical model, such as the HARMONIE-ALADIN \cite{bengtsson2017harmonie,aladin} model. These models have as input the low resolution ground truth plus the additional information necessary to compute the downscaling result. Thus, if we represent, coherently with the Super-resolution problem, the high resolution image as $x$, the low resolution images as $y$, the statistical downscaling model as $H\!A$ and the additional information as $\rho_{H\!A}$, we can define the downscaling process as: 
\begin{equation}
        x = H\!A(y,\rho_{H\!A})
\end{equation}
We observe that this equation is closely related to Equation \ref{eq:inverse_problem}, which establishes a relationship between high-resolution and low-resolution images. This similarity effectively equates the task of super-resolution in image processing to the problem of statistical downscaling in meteorological studies.

The primary distinction between the two problems lies in the inverse nature of their low-resolution/high-resolution relationship. Contrary to the super-resolution process described in Equation \ref{eq:conv_degradation}, where the low-resolution image is derived from an unknown high-resolution counterpart, in statistical downscaling, the situation is reversed. Here, the high-resolution image is an enhanced version of the low-resolution image.

We can thus define the new super-resolution task as the problem of approximating the behaviour of the $H\!A$ model. Considering that our goal is to proceed in a data driven manner and to avoid the utilization of additional information for the downscaling process, we define our approximate model without the use of $\rho_{H\!A}$.

\begin{equation}
        \hat{x} = \Phi(y) \, ,  \: \hat{x} \approx x
\end{equation}
Where $\Phi$ is our super-resolution model and $\hat{x}$ is the approximation of the ground truth $x$ generated by the statistical downscaling model.

\subsection{Super-Resolution Methods}
Traditional methods of Super-resolution typically involved interpolation techniques like bicubic or bilinear interpolation, which enlarged images by inserting new pixels based on the color values of neighboring pixels. These methods, while straightforward and computationally inexpensive, often led underwhelming results, especially when dealing with significant upsampling. 

In recent years the research has instead focused on deep learning approaches \cite{7115171, dong2014learning, wang2015, li2019feedback, dai2019second, hu2019meta, shi2016real, lu2022transformer}. Deep learning has revolutionized the field of super-resolution by substantially enhancing the quality and effectiveness of image upsampling. This advancement includes not only improved image clarity but also the capability to learn from data and perform context-aware upsampling. Central to this advancement are Convolutional Neural Networks (CNNs), which form the core of many state-of-the-art super-resolution methods, with architectures like SRCNN \cite{dong2014learning} and VDSR \cite{kim2016accurate} leading the way. 

Subsequent improvements to the CNN architecture include ESPCN \cite{shi2016real}, which is designed to operate in real-time both for images and videos, implementing a sub-pixel convolution layer at the very end of the network to aggregate LR feature maps and simultaneously perform projection to high-dimensional space to reconstruct the HR image. 
Subsequent advancements have been significantly enhanced by the introduction of residual networks. These networks incorporate skip connections in their design, effectively addressing the issue of vanishing gradients and enabling the construction of deeper networks, which in turn boosts overall performance. Example of residual networks for SR include Enhanced Deep Super-Resolution (EDSR) \cite{lim2017enhanced}, which modifies the ResNet architecture \cite{he2016deep} proposed originally for image classification to work with the SR task, and CARN \cite{ahn2018fast}, which utilize residual blocks in intermediate layers which are then cascaded and converged onto a 1×1 convolutional layer. 

The research in super-resolution methods based on CNN architectures has seen a remarkable diversification in the following years, novel architectural innovations include the use of attention mechanism \cite{zhang2018image, choi2017deep}, multi-branch designs \cite{ren2017image,hu2018single} and dense connections \cite{haris2018deep, tong2017image}.

In more recent years, the focus of research in super-resolution has undergone a significant paradigm shift, moving away from traditional CNN-based predictive models towards more advanced generative approaches. The first steps in this direction were based around the Generative Adversarial Network (GAN) architecture \cite{goodfellow2020generative}, which employs a game-like approach, with the model being comprised of two competing elements, the generator and the discriminator. The generator is trained to produce a HR image in such a way that the discriminator cannot distinguish it from the original HR image. Training in this adversarial manner, HR images with better perceptual quality are generated. The most notable applications of GAN to the super-resolution task include SRGAN \cite{ledig2017photo} and ESRGAN \cite{wang2018esrgan}.

The state of the art in super-resolution has since shifted from GAN-based approaches to diffusion and transformer based approaches. Transformers \cite{vaswani2017attention} are the most relevant architecture for natural language processing. Recently, their adaptation to the visual domain, known as visual transformers, has also begun to set new benchmarks, establishing themselves as state-of-the-art for various visual tasks. Transformers excel in handling sequential data. They utilize self-attention to process input data in a way that each element of the sequence can directly attend to every other element, enabling the model to capture complex relationships. In the case of images, the process involves treating an image as a sequence of patches or pixels. Thanks to the ability to learn contextual relationships transformer based models such as SWINIR \cite{liang2021swinir} and SWIN2SR \cite{conde2022swin2sr} are at the forefront of the research in super-resolution. 

Similarly to transformers, diffusion models \cite{song2020denoising} have recently demonstrated the capability to achieve state of the art performance in super resolution tasks \cite{li2022srdiff}. Diffusion models are a type of generative model that simulate a process similar to heat diffusion. They start by gradually adding noise to an image (or any data) until it turns into random noise. The model then learn to reverse this process, reconstructing the original data from the noisy version. This distinctive capability positions diffusion models at the forefront of generative tasks, avoiding the training instability that can affects GAN and Trasformers. In this work, we focus on the Denoising Diffusion Implicit Model (DDIM) as our primary architecture.

\subsection{Metrics}
In our research, we will evaluate the quality of our super-resolution methodologies using three established metrics: Mean Squared Error (MSE), Peak Signal-to-Noise Ratio (PSNR), and Structural Similarity Index Measure (SSIM). 
Mean Squared Error (MSE) is a widely used metric in a wide range of applications, including super-resolution. It quantifies the difference between the predicted values and the actual values by averaging the squares of the errors. The error in this context is the difference between each predicted value and its corresponding true value.

\begin{equation}
    MSE = \frac{1}{n} \sum_{i=1}^{n} (y_i - \hat{y}_i)^2
\end{equation}

Peak Signal-to-Noise Ratio (PSNR) is a prominent metric used in the field of image processing to evaluate the quality of reconstructed, compressed, or denoised images in comparison to the original image. It is defined as the ratio between the maximum possible power of a signal (represented by pixel values in an image) and the power of corrupting noise that affects its fidelity. 

\begin{equation}
    PSNR = 20 \cdot \log_{10}\left(\frac{1}{\sqrt{MSE}}\right)
\end{equation}

Finally, Structural Similarity Index (SSIM) is an advanced metric used for measuring the similarity between two images. Developed to provide a more accurate and perceptually relevant assessment of image quality, SSIM differs from traditional metrics like MSE and PSNR by considering changes in structural information, luminance, and contrast. Unlike MSE and PSNR, which focus on pixel-to-pixel differences, SSIM evaluates changes in structural patterns, making it more aligned with human visual perception. SSIM can be written as: 

\begin{equation}
    SSIM(x, y) = \frac{(2 \mu_x \mu_y + C_1)(2 \sigma_{xy} + C_2)}{(\mu_x^2 + \mu_y^2 + C_1)(\sigma_x^2 + \sigma_y^2 + C_2)}
\end{equation}
where the metric is computed between two corresponding patches represented by $x$ and $y$, with $\mu_x$ and $\mu_y$ being the respective pixel average, $\sigma_x^2$ and $\sigma_y^2$ the variance and $\sigma_{xy}$ the covariance. $C_1$ and $C_2$ are two custom parameters to o stabilize the division with weak denominator.

\subsection{Diffusion Models}

Diffusion models are a class of probabilistic generative models that are particularly effective in modeling complex, high-dimensional data distributions. At the core of diffusion models lies the mathematical concept of a diffusion process, that is, a stochastic process that describes the continuous random movement of particles over time, modeling the spread or diffusion of some quantity in space or time, where the particles tend to move from regions of high concentration to regions of low concentration, resulting in a gradual blending or mixing of the quantity. In the context of machine learning, diffusion models leverage the principles of diffusion processes to model the generation of data. Instead of directly sampling data points from a fixed distribution, these models iteratively transform a simple initial distribution, typically a known distribution like a Gaussian or uniform distribution, into the desired complex data distribution. The main idea is to perform a series of diffusion steps, where each step updates the probability distribution of the data. This is achieved by adding Gaussian noise to the current data samples and iteratively refining them. 

From a mathematical perspective, considering a distribution $q(x_0)$ which generates the data, generative models aim to find a parameter vector $\theta$ such that the distribution $p_\theta(x_0)$ parameterized by a neural network approximates $q(x_0)$. 

Denoising Diffusion Probabilistic Models (DDPM)\cite{DDPM} assume the generative distribution $p_\theta(x_0)$ to have the form 
\begin{equation}
p_\theta(x_0)) = \int p_\theta(x_{0:T})dx_{1:T}
\end{equation}
given a time range horizon $T>0$.
where the Markov Chain formulation is : 
\begin{equation}
p_\theta(x_{0:T})=p_\theta(x_T)\prod_{t=1}^{T}p_\theta(x_{t-1}\vert x_t).
\end{equation}

Training is traditionally based on a variational lower bound of the negative loglikelihood. Considering that the Kullback-Leibler Divergence $D_\text{KL} = D_\text{KL}(q(x_{1:T}\vert x_0) \| p_\theta(x_{1:T}\vert x_0) )$ is positive, we obtain: 
\begin{equation}
- \log p_\theta(x_0) \leq - \log p_\theta(x_0) \!+\! D_\text{KL}
\end{equation}
We can thus expand the second term to derive the training loss $L_\theta$:
\begin{align}
& \hspace{.4cm}= -\log p_\theta(x_0) \! +\! \mathbb{E}_q \Big[ \log\frac{q(x_{1:T}\vert x_0)}{p_\theta(x_{0:T}) / p_\theta(x_0)} \Big] \nonumber \\
& \hspace{.4cm}= -\log p_\theta(x_0) \!+\! \mathbb{E}_q \Big[ \log\frac{q(x_{1:T}\vert x_0)}{p_\theta(x_{0:T})} + \log p_\theta(x_0) \Big] \nonumber \\
& \hspace{.4cm}= \mathbb{E}_q \Big[ \log q(x_{1:T}\vert x_0) -p_\theta(x_{0:T}) \Big] = L_\theta \label{eq:loss_elbo}
\end{align}

In the case of diffusion models, the latent space has typically the same dimension of
the visible space. 

The inference procedure $q(x_{1:T} \vert x_0)$ is fixed, in contrast with different 
latent variables models like Variational Autoencoders (VAEs) \cite{RezendeMW14,VAEKingma,VAEGreen} where it is learned along with the generative distribution.

In the particular case of Denoising Diffusion Implicit Models (DDIMs)\cite{2020arXiv201002502S}, used
in this work, the authors considered a non-Markovian  diffusion process
\begin{equation}
    q_\sigma (x_{1:T} \vert x_0) = q_\sigma(x_T \vert x_0) \prod_{t=2}^T q_\sigma (x_{t-1} \vert x_t, x_0)
\end{equation}
where $q_\sigma(x_T \vert x_0) = \mathcal{N}(x_T \vert \sqrt{\alpha_T} x_0, (1 - \alpha_T) \cdot I)$, and
\begin{align}\label{eq:non_markovian_reverse_diffusion}
    q_\sigma (x_{t-1} \vert x_t, x_0) = \mathcal{N} \Bigl( x_{t-1} \Big \vert \mu_{\sigma_t}(x_0, \alpha_{t-1}); \sigma_t^2 \cdot I \Bigr)
\end{align}
with 
\[
\begin{array}{l}
    \mu_{\sigma_t}(x_0, \alpha_{t-1}) = \\
    \hspace{.5cm}\sqrt{\alpha_{t-1}} x_0 + \sqrt{1 - \alpha_{t-1} - \sigma_t^2} \cdot \frac{x_t - \sqrt{\alpha_t} x_0}{\sqrt{1 - \alpha_t}}.
\end{array}
\]
The definition of $q(x_{t-1} \vert x_t,x_0)$ is cleverly chosen in order to ensure two important aspects of the diffusion process of DDPM: the Gaussian nature of $q(x_{t-1} \vert x_t,x_0)$ and the fact that the marginal distribution $q_\sigma(x_t\vert x_0) = \mathcal{N}(x_t \vert \sqrt{\alpha_t} x_0; (1 - \alpha_t) \cdot I)$ is the same as in DDPM. 
Thanks to the latter property, $x_t$ can be expressed 
as a linear combination of $x_0$ and a noise variable $\epsilon_t \sim \mathcal{N}(\epsilon_t \vert 0; I)$:
\begin{align}\label{eq:xt_from_x0}
    x_t = \sqrt{\alpha_t} x_0 + \sqrt{1 - \alpha_t} \epsilon_t.
\end{align}
The next step consists in defining a trainable generative process $p_\theta(x_{0:T})$ where  $p_\theta(x_{t-1}\vert x_t) $ leverages the structure of $q_\sigma(x_{t-1} \vert x_t, x_0)$. The idea is that given a noisy observation $x_t$, one starts making a prediction
of $x_0$, and then use it to obtain $x_{t-1}$ 
according to equation~\ref{eq:non_markovian_reverse_diffusion}.

In practice, we train a neural network $\epsilon_\theta^{(t)}(x_t, \alpha_t)$ to map a given $x_t$ and a noise rate $\alpha_t$ to an estimate of the noise $\epsilon_t$ added to $x_0$ to construct $x_{t}$. As a consequence, $p_\theta(x_{t-1} \vert x_t)$ becomes a $\delta_{f_\theta^{(t)}}$, where
\begin{align}\label{eq:approx_DDIM}
    f_\theta^{(t)}(x_t, \alpha_t) = \frac{x_t - \sqrt{1 - \alpha_t} \epsilon_\theta(x_t, \alpha_t)}{\sqrt{\alpha_t}}.
\end{align}
Using $f_\theta^{(t)}(x_t, \alpha_t)$ as an approximation of 
$x_0$ at timestep $t$, $x_{t-1}$ is then obtained as follows:
\begin{align}
x_{t-1} = & \sqrt{\alpha_{t-1}}\cdot f_\theta^{(t)}(x_t, \alpha_t) + \nonumber \\ 
& 
\sqrt{1-\alpha_{t-1}-\sigma_t^2}\cdot \epsilon_\theta(x_t, \alpha_t) \label{eq:x_t_minus_1}
\end{align}.

As for the loss function, the term in Equation \ref{eq:loss_elbo}
can be further refined expressing $L_\theta$ as the sum of the following terms \cite{Sohl-DicksteinW15}:
\begin{equation} 
\label{eq:loss_sum}
L_\theta = L_T + L_{t-1} + \dots + L_0 
\end{equation}
where
\[
\begin{array}{l}
L_T = D_\text{KL}(q(x_T \vert x_0) \parallel p_\theta(x_T)) \\
L_t = D_\text{KL}(q(x_t \vert x_{t+1}, x_0) \| p_\theta(x_t \vert x_{t+1})) \\
\hspace{3.5cm}\text{ for }1 \leq t \leq T-1 \\
L_0 = - \log p_\theta(x_0 \vert x_1)
\end{array}
\]
All previous distributions are Gaussian and their KL divergences 
can be calculated in closed form using, obtaining the following formulation:
\begin{equation}
L_t = \mathbb{E}_{t \sim [1, T], x_0, \epsilon_t} \Big[\gamma_t\| \epsilon_t - \epsilon_\theta(x_t, t) \|^2 \Big]
\end{equation}
This can be simply interpreted as the weighted Mean Squared Error between the predicted and the actual noise a time $t$. 

Due to experimental evidence, the weighting parameters are usually ignored in practice, since 
the training process works better without them. 

The pseudocode for training and sampling is given in the following Algorithms. 

\begin{algorithm}[H]
    \centering
    \caption{Training\label{algorithm1}}
    \begin{algorithmic}[1]
        \Repeat
        \State $x_0 \sim q(x_0) $  
        \State $t \sim $Uniform({1,..,T}) 
        \State $\epsilon \sim \mathcal{N}(0;I)$ 
        \State $x_t = \sqrt{\alpha_t} x_b + \sqrt{1\!-\! \alpha_t} \epsilon$  
        \State Backpropagate on
        $\lvert \lvert \epsilon - \epsilon_{\theta} (x_t, \alpha_t )\rvert \rvert^2$ 
        \Until converged
    \end{algorithmic}
\end{algorithm}

Sampling is an iterative process, starting from a purely noisy image $x_T \sim \mathcal{N}(0,I)$. The denoised version of the image at timestep $t$ is obtained using equation~\ref{eq:x_t_minus_1}.

\begin{algorithm}[H]
    \centering
    \caption{Sampling}\label{algorithm2}
    \begin{algorithmic}[1]
        \State $x_T \sim \mathcal{N}(0,I)$
        \For {$t = T,...,1$}
       \State $\epsilon = \epsilon_\theta(x_a,x_t,\alpha_t)$ 
       \State $\tilde{x}_0 = \frac{1}{\sqrt{\alpha_t}} (x_t - \frac{1-\alpha_t}{\sqrt{1 - \alpha_t}} \epsilon)$ 
       \State $x_{t-1} = \sqrt{\alpha_{t-1}}\tilde{x}_0 + \sqrt{1 - \alpha_{t-1}}\epsilon$  
        \EndFor
    \end{algorithmic}
\end{algorithm}
\section{Related Works}
The existing literature primarily focuses on two key areas: the application of super-resolution techniques to downscale weather variables and the assessment of regional high-resolution reanalysis datasets through comparison with in-situ observations.
In \cite{Ji2020Downscaling} the authors effectively utilize super-resolution neural techniques to perform downscaling of precipitation data. The study introduces and evaluates three super-resolution deep learning frameworks: the Super-Resolution Convolutional Neural Network (SRCNN), Super-Resolution Generative Adversarial Networks (SRGAN), and Enhanced Deep Residual Networks for Super-Resolution (EDSR). These frameworks are applied to refine the spatial resolution of daily precipitation forecasts in Southeast China. Specifically, they enhance the resolution from approximately 50 kilometers to finer scales of about 25 kilometers and 12.5 kilometers. For comparison, Bias Correction Spatial Disaggregation (BCSD) as a traditional SD method is also performed under the same framework. The precipitation forecast data utilized in this research are sourced from various Ensemble Prediction Systems (EPSs), including the European Centre for Medium-Range Weather Forecasts (ECMWF), the National Centers for Environmental Prediction (NCEP), and the Japan Meteorological Agency (JMA). The results demonstrate that all three SR models effectively capture intricate local details in precipitation patterns with EDSR achieving the best overall performance. 
In \cite{nguyen2023climatelearn} the authors propose a climate research library presenting a series of tasks, including statistical downscaling. They choose two different settings, the first is downscaling 5.625$^{\circ}$ ERA5 to 2.8125$^{\circ}$ ERA5 with hourly intervals and at global scale, using as target variables Z500, T850 and T2m. In the second setting the authors consider downscaling from 2.8125 $^{\circ}$ ERA5 to 0.75$^{\circ}$ PRISM over the continental United States at hourly intervals for the variable daily max T2m. 
With PRISM being a dataset of interpolated in-situ observations, this setting is particularly interesting as it performs downscaling from a reanalysis to an observational dataset. For both settings the training period is from 1981 to 2016, and the testing period is the year 2017. The authors tests a series of model, including ResNet, U-Net, and ViT, nearest and bilinear interpolation. 
In \cite{Harder2022Generating} the authors perform statistical downscaling with super-resolution neural models, while guaranteeing physical constraints via the use of hard constraining layers enforcing conservation of mass. In the experimental setting the ERA5 dataset was used, focusing on the total column water (tcw) variable. The super-resolution setting was achieved by downsampling ERA5 by factors ranging from 2 to 16. The obtain results reveal increased performance with respect of non-constrained models.  
In \cite{PELOSI2023108556} the authors analyze the performance of ERA5, ERA5-Land and CERRA on the territory of Sicily in estimating air temperature, air relative humidity, solar radiation, and wind speed. The estimated weather variables are then compared to in-situ measurement,  with the outcomes revealing that CERRA provides the best estimations overall. 
In \cite{monteiro2023multi} the authors perform an in-situ validation for various modeling systems of global and regional reanalyses including ERA5, ERA5-Land, ERA5-Crocus, CERRA-Land, UERRA MESCAN-SURFEX and MTMSI. The validation data included observational references (in situ, gridded observational datasets and satellite observations) across the European Alps from 1950 to 2020 for a set of weather variables including snow cover and its first-order drivers, temperature and precipitation. Their results reveal that no modeling strategy outperforms all others within their experimental sample.
In \cite{environsciproc2023026111} the authors assess the accuracy of CERRA for human-biometeorological and heat–health applications on the territory of Greece, taking into consideration temperature and wind speed. The findings indicate that CERRA outperformed in the majority of statistical assessments and demonstrated performance on par with ERA5-Land in the remaining ones.
The study carried out by \cite{cavalleri2023inter} examines the performance of various reanalysis datasets, including ERA5, ERA5-Land, CERRA, MERIDA, VHR-REA\_IT, in accurately reproducing the temperature patterns characteristic of the Italian peninsula. Although the research is not yet completed, Initial results reveal that the utilized reanalyses effectively replicate the temperature climatology across Italy for a range of spatial resolutions, from 31 km to 2.2 km, exhibiting only minimal degradation in performance when modeling smaller-scale processes.


\section{Datasets}
\label{sec:dataset}

In our study, we utilized three distinct datasets to perform super-resolution and validation. We employed the ERA5 \cite{Hersbach_ERA5_2018}  dataset as our low-resolution baseline, characterized by its spatial precision of around 30 km, and compared it with the high-resolution CERRA \cite{Schimanke_CERRA_2021} dataset, having a finer spatial resolution of around 5.5 km over the European region. For real-world validation of our results, we used the IGRA V2 \cite{durre2018enhancing} in-situ dataset.

Both ERA5 and CERRA are cutting-edge in the field of numerical weather reanalysis, widely utilized in a spectrum of applications including forecasting, predicting extreme weather events, and conducting climate studies. While ERA5 offers a global perspective, CERRA provides a more region-specific focus, covering only Europe. Our selection of datasets was strategic; CERRA uses ERA5 as a boundary condition, aligning well with our super-resolution objectives where a robust correlation exists between low-resolution and high-resolution images.

Notably, CERRA adopts a conical projection optimal for high latitudes, in contrast to ERA5's cylindrical projection. This necessitated a conversion for direct comparison. Focusing on the Italian peninsula, an area with relatively lower latitudes, we conducted our comparative analysis using the cylindrical projection. In this converted format, the reference area for CERRA amounted to 256x256 pixels, while for ERA5, it was 52x52 pixels.

For this comparative study, we chose wind speed as our primary variable, offering a valid example through which examine the downscaling performance. 

\subsection{ERA5}
ERA5 \cite{Hersbach_ERA5_2018}, short for the Fifth Generation of the European Centre for Medium-Range Weather Forecasts (ECMWF) Reanalysis, is a well known and vastly utilized dataset in the realm of climate and atmospheric research. ERA5, encompassing a period from 1940 to the present, provide a detailed portrait of the Earth's atmospheric system. It provides hourly estimates of a large number of atmospheric, land and oceanic climate variables with an horizontal resolution of 0.25 degrees (around 30km) and 137 levels in the vertical from the surface to 0.01 hPa (about 80km). An innovative aspect of The ERA5 reanalyses is its timely availability.  Collecting the necessary data and their computation results in a production latency of approximately 5 days. This timely availability makes ERA5 a valuable tool for studying events in the recent past and forecasting future weather.

The assimilation system used in ERA5, based on the ECMWF Integrated Forecasting System (IFS), incorporates advancements in numerical weather prediction models, radiative transfer schemes, and satellite data utilization, contributing to the dataset's reliability and precision \cite{hersbach2020era5}. 

ERA5 is a state-of-the-art global atmospheric reanalysis and it serves as a robust tool for understanding and analyzing the Earth's climate system, offering a global representation of atmospheric, land, and oceanic conditions spanning several decades. ERA5 over the years it has been a crucial asset for scientists from a variety of fields, with the dataset appearing in a vast amount of studies. Its applications span several key areas:

\begin{itemize}
    \item Analysis of Precipitation trends: ERA5 data is extensively used for tracking precipitation changes, as evidenced in multiple recent works \cite{ yuan2021feasibility, lavers2022evaluation, chiaravalloti2022long, shen2022changes}. These works highlight its utility in observing and understanding changes in precipitation patterns relative to climate change.
    \item Investigation of Temperature Trends: The dataset is instrumental in analyzing long-term temperature trends, with studies ranging from global temperatures \cite{yilmaz2023accuracy,liu2020global} to on urban heat \cite{lee2024improved}.
    \item Wind Analysis: Recent works demonstrates its application in studying wind patterns for renewable energy monitoring \cite{olauson2018era5} and addressing climate changes \cite{cai2021wind}.
    \item Extreme Events Study: ERA5's role is pivotal in advancing our understanding of extreme climate events, exploring storm surge \cite{dullaart2020advancing} and heatwaves \cite{al2024assessing} and extreme precipitations \cite{dong2021evaluation, wei2023spatiotemporal}.
\end{itemize}

The high temporal resolution of ERA5 is particularly valuable in detecting shifts in climate patterns, which is crucial for understanding global warming and climate extremes. This information enables researchers to refine climate models, project future scenarios, and assess the potential impacts of climate change on both regional and global scales.

Additionally, ERA5 data finds important applications in other fields such as agriculture \cite{Rolle2020Improved}, water resources management \cite{Zhang2021Assessment, Tarek2020Evaluation}, and urban planning and infrastructure design \cite{Jiao2021Evaluation}.
In summary, ERA5 represents an invaluable asset in weather and climate studies, playing a pivotal role in ongoing scientific research.

\subsection{CERRA}
The Copernicus Regional Reanalysis for Europe (CERRA) is a sophisticated high-resolution regional reanalysis (RRA) dataset specifically designed for the European region. It is a product of the European Copernicus program, executed through a contract with the Swedish Meteorological and Hydrological Institute (SMHI), in collaboration with subcontractors Meteo-France and the Norwegian Meteorological Institute. CERRA offers data at a fine 5.5km horizontal resolution.

\begin{figure}[h]
\centering
\includegraphics[width=\linewidth]{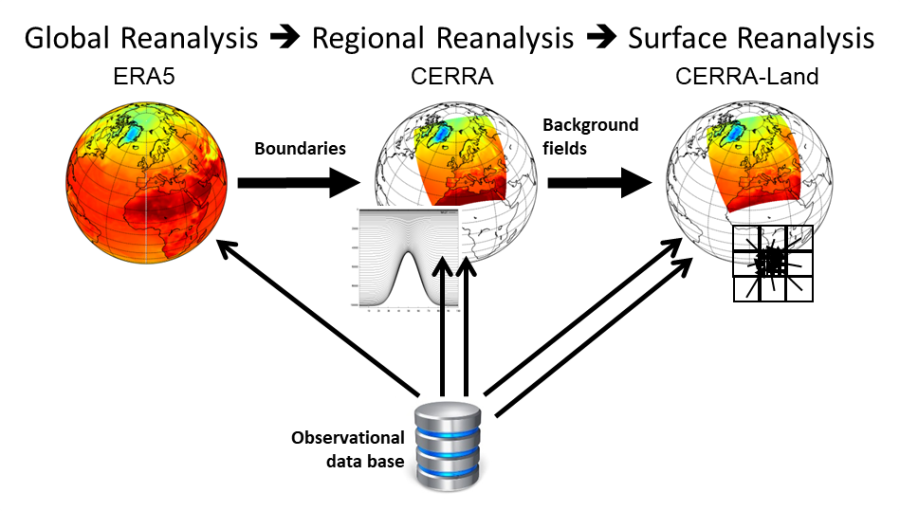}
\caption{Three different stages of reanalysis: (left) the Global Reanalysis ERA5 will be used as boundary condition, (middle) CERRA Regional Reanalysis, and (right) a CERRA 2D Reanalysis for the near surface. The arrows indicate that the amount of observational information used for the reanalysis progressively increases from the global to the regional reanalysis. Source: Copernicus-ECMWF CERRA Presentation, 2024 \cite{presentationCerra}}
\label{fig:cerra_recap}
\end{figure}

This dataset achieves its detailed resolution by utilizing the global ERA5 reanalysis dataset, which provides both initial and boundary conditions. In addition to inputs from ERA5, CERRA's regional reanalysis incorporates higher-resolution observational data and physiographic datasets that describe surface characteristics. This comprehensive approach is illustrated in Figure \ref{fig:cerra_recap}.

CERRA's output includes both forecasts and reanalyses. Weather forecasting in this context involves analyzing the current state of the atmosphere and terrestrial and marine surfaces. These forecasts are generated using mathematical and physical numerical models that start from this analysis. During the reanalysis phase, a weather forecasting model is initially used to estimate the atmospheric state at a specific time. This 'first guess' is then refined through a process known as 'data assimilation', which involves correcting the model based on observational data. This process ensures that the reanalysis provides an accurate historical record of weather conditions.

CERRA contains a wide range of meteorological variables, including temperature, humidity, wind speed and direction, precipitation, and cloud cover. The data is available on a regular grid, while the availability in terms of temporal resolution follows a different structure with respect to the hourly one present in ERA5, differentiating between reanalysis and forecast time. For each day 8 reanalysis are available, starting from 00 UTC every three hours until 21 UTC. The prevailing assumption is that analysis data for these time periods exhibits superior quality compared to forecasts for the same time frames, as it is inherently more aligned with observations. Starting from each reanalysis data available, the next 6 hours are processed through forecasting. And for reanalyses starting at 00 UTC and 12 UTC the next 30 hours are processed, with a 3 hour interval after the first 6 hours. In this way users can select, even for the same hour, the forecast they prefer, or if available, the reanalysis result. However, this means that even if CERRA provides data for each hour, only the 8 present in the reanalysis schedules guarantee a corrective step using observations, while the others are forecast results.

The CERRA dataset is a valuable resource for a wide range of applications, including the study of climate trends, renewable energy forecasting, water resource management, and risk assessment. For climate research, CERRA provides a historical reconstruction of meteorological variables, which is essential for understanding long-term climate trends and the impacts of climate change in Europe. A study \cite{galanaki2023validating} demonstrated that CERRA outperforms the ERA5-Land reanalysis in replicating temperature and other biometeorological variables, making it particularly useful for assessing heat-related health risks. In the field of renewable energy, CERRA's accurate weather observations are crucial for developing forecasting tools for wind and solar power, aiding the integration of these renewable sources into the power grid \cite{nicodemo2023remote}. Water resource management also benefits from CERRA's ability to represent precipitation \cite{Verrelle2021CERRA-Land} and snow \cite{environsciproc2023026016}, helping to manage resources and assess the risks of drought, flooding and economic risks. Lastly, CERRA has been demonstrated as a viable substitute for ground-based agrometeorological measurements, effectively capturing data on air temperature, actual vapor pressure, wind speed, and solar radiation, thus making CERRA a valuable tool for regional studies in water resource management \cite{PELOSI2023108556}. These applications underscore the importance of the CERRA dataset in supporting adaptation actions, policy development, and climate monitoring and research across Europe.

Given the importance of CERRA in a variety of applications, it would be ideal to have a real-time production of this data, to effectively study recent events. Unfortunately delays in the production and data gathering slowed the release of CERRA, with a delay of more than 2 years at the time of writing. 

\subsection{IGRA V2}
For the validation task, data from the Integrated Global Radiosounding Archive (IGRA) V2 \cite{durre2016igra} dataset has been employed. IGRA is a comprehensive collection of global radiosounding profiles, encompassing temperature, humidity and wind measurements and it is freely accessible.

The in-situ data used refer to meteorological stations located in the same range of longitude and latitude represented by the other two datasets. The result of this selection is 14 stations based spread on the selected area.

IGRA V2 undergoes validation through various processes to ensure data quality and reliability. The validation includes quality control of radiosonde observations of temperature, humidity, and wind at stations. The data is required to have adequate documentation of its digital format, and only observed values are included, while estimated values are not, so this dataset represents a good source of information to validate results for the field of reanalysis and downscaling.

\section{Methodology}
In this section, we will provide a detailed exposition of our diffusion technique applied to super-resolution in statistical downscaling, which includes a comprehensive validation component. Our discussion begins with an overview of the preliminary steps required to prepare our case study, including the harmonization of features from various datasets. We then offer a detailed exploration of our primary and most effective model, the diffusion model. Following this, we compare the diffusion model with other super-resolution approaches to highlight relative strengths and weaknesses. Next, we outline the validation methods employed to ensure the accuracy of our model's output against real-world measurements. Lastly, we evaluate our results and provide an in-depth analysis of our findings.

\begin{figure}[h]
\centering
\includegraphics[width=\linewidth]{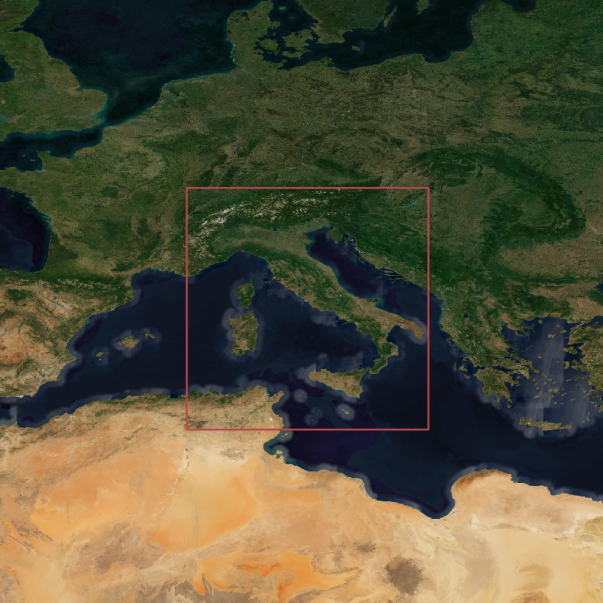}
\caption{Selected area for our study, delimited by the following geographical coordinates: North at 47.75°, South at 35°, East at 18.75°, and West at 6°.}
\label{fig:map1}
\end{figure}

\subsection{Data Selection \& Preprocessing}
\label{sec:preprocessing}

In this section we analyze the essential preprocessing steps required for defining the downscaling process from ERA5 to CERRA. Both ERA5 \cite{Hersbach_ERA5_2018} and CERRA \cite{Schimanke_CERRA_2021} are freely available from the Copernicus Program \cite{C3S_ERA5_2023} website. 

In our study, we have chosen to concentrate on the meteorological variable of wind speed. This decision was influenced by several factors. Firstly, we aimed to simplify our experimental framework by focusing on a single variable. We selected wind speed because, compared to variables like rainfall, it is more physically self-contained and thus easier to analyze in isolation. Additionally, wind speed holds significant societal relevance, particularly in the context of extreme weather events and its implications for energy generation in wind farms. 

We utilized wind speed at 10 meters above the surface. In the case of CERRA wind speed is already included as a main variable, while for ERA5 we computed it from both the zonal $u$ and the meridional $v$ wind components by $ \sqrt{u^2 + v^2}$. The utilized values are instantaneous.

For our analysis, we extracted a specific region from the ERA5 dataset, delimited by the following geographical coordinates: North at 47.75°, South at 35°, East at 18.75°, and West at 6°. This targeted extraction ensures a focused study area for our research. This area coincides with a square containing Italy, Switzerland, Slovenia and partially Tunisia, Algeria, France, Austria, Croatia, Hungary, Bosnia \& Erzegovina and Montenegro, as shown in Figure \ref{fig:map1}. 

As mentioned in Section \ref{sec:dataset}, the two datasets have a different projection mechanism, with ERA5 using a cylindrical projection and CERRA a Lambert Conformal Conic projection.  
Given that our study area is located at mid latitudes, where distortions in cylindrical projections are relatively minimal, we chose to equalise the two projections by re-projecting CERRA using a cylindrical projection consistent with ERA5. The projection was realized utilizing the CDO \cite{schulzweida_2023_10020800} library.

CERRA provides data samples with a 3-hour temporal resolution, whereas ERA5 offers an hourly resolution. To align these datasets, we extracted ERA5 measurements corresponding to the times 00:00, 03:00, 06:00, 09:00, 12:00, 15:00, 18:00, and 21:00, matching CERRA's schedule. Our training set encompasses data from 2010 to 2019. For testing purposes, we selected the years 2020 and 2009.  The choice of non-contiguous test years is motivated by aiming to widen our evaluation range and the availability of in-situ observations for validation. 

The training sets for both ERA5 and CERRA have been normalized to a range between 0 and 1, with each entry being divided by the maximum value within its respective dataset. Similarly, the test sets for each dataset have been normalized using the same maximum value as determined from their corresponding training sets.

\begin{figure}[h]
\centering
\includegraphics[width=\linewidth]{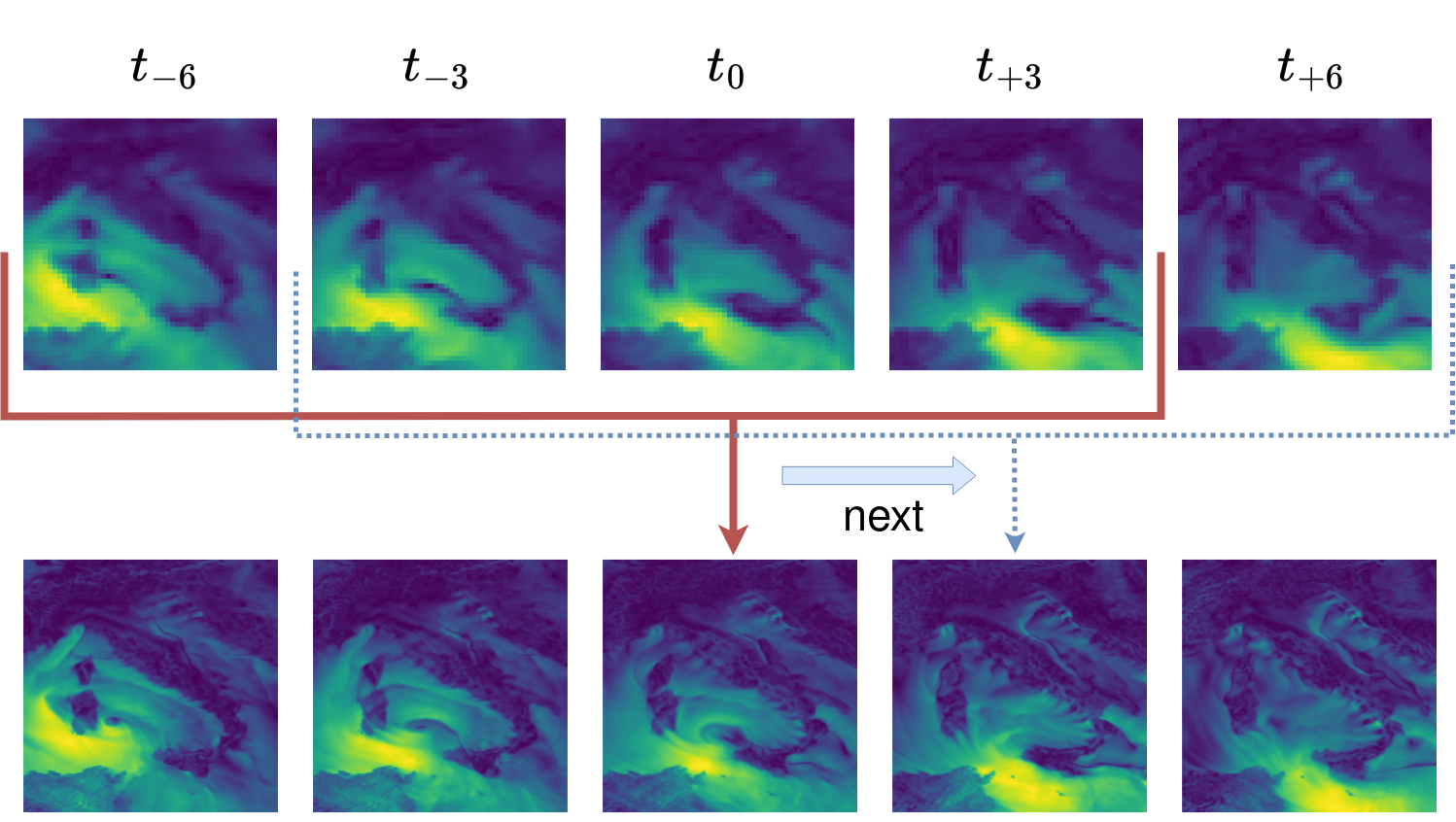}
\caption{The downscaling diffusion process utilize as input a set of low resolution images. For predicting the high resolution image at time $t_0$ the conditioning information include the low resolution images at times $t_{-6}$ , $t_{-3}$, $t_0$, $t_{+3}$ where each step is 3 hours long. }
\label{fig:sequenceto1}
\end{figure}

\subsection{Diffusion Model Architecture}
In this section, we describe our super-resolution diffusion model. This includes an examination of the super-resolution (SR) framework, the specifics of the conditioning process, ensemble methods, and various implementation details.

\subsubsection{SR framework}
Super-resolution for neural networks implies the use of some form of upsampling, which may be implemented in different ways and in different positions within the network. Generally speaking, we can define two frameworks for SR: pre-upsampling, when the upsampling is done in the first part of the network, and post-upsampling, when the upsampling part is implemented at the very end of the network. If the upsampling is performed at the start, the spatial dimension of the features is increased, usually granting the network better overall performance at the expense of computational complexity. The first framework is generally used in setting where the main target is performance, such as SRGAN \cite{wang2018esrgan}, the latter is more common in performance-oriented model such as ESPCN \cite{shi2016real}. For our diffusion model we opted for a pre-upsampling framework. 

The process of upsampling can be executed using various methods. Some methods employ learnable parameters, such as transposed convolution, while others, like 2D upsampling or bilinear and bicubic upsampling, do not use learnable parameters. In our model we opted for preemptive bilinear upsampling, due to its practicality. Additionally, our experiments with methods involving upsampling with learnable parameters yielded similar results.

\begin{figure*}[h]
\centering
\includegraphics[width=\linewidth]{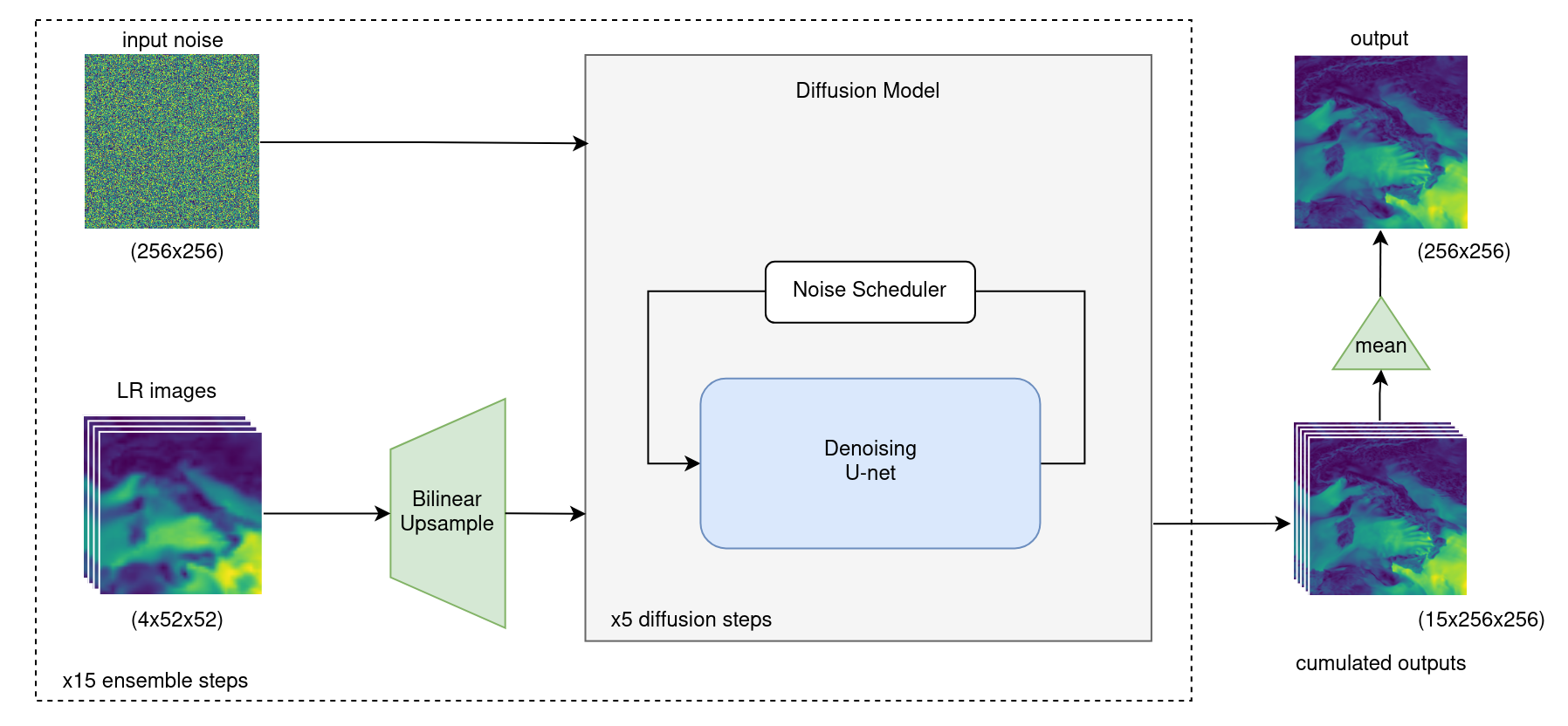}
\caption{The super-resolution diffusion framework employs a sequence-to-one strategy, where pre-upsampling is achieved via bilinear interpolation. The diffusion process is comprised of five diffusion steps, using a denoising U-net to progressively denoise the input. A noise scheduler orchestrates this process. This procedure is replicated 15 times, generating an ensemble of potential super-resolution outcomes. These outcomes are then amalgamated using a mean operation to produce the final output.}
\label{fig:big_diffusion}
\end{figure*}

For what concerns the super resolution task setting, we decide to utilize a sequence-to-one approach, as represented in Figure \ref{fig:sequenceto1} . In this approach the super resolution model is trained to produce the high resolution images while having access to multiple timestamps of the low resolution counterpart. Specifically, experimental results showed that the optimal sets of conditioning information for the prediction of the high resolution image at time $t_0$ contains the low resolution images at hours $t_{-6}$, $t_{-3}$, $t_0$, $t_{+3}$. In our implementation the main bulk of our data is kept on disk while a data generator feeds the model the sequences for both training the testing operations, with the first being a random set of sequences and the second a contiguous set. 

\subsubsection{Denoising and conditioning}

Diffusion models essentially operate as iterative denoising algorithms. Their main trainable component is the denoising network, denoted as $\epsilon_\theta(x_t, \alpha_t)$. This network receives as input the noisy input, $x_t$, and a corresponding noise variance, $\alpha_t$, with the objective of estimating the amount of noise infiltrating the image. The training of this underlying denoising network is done conventionally. An initial sample, $x_0$, is extracted from the dataset and subjected to a predefined amount of random noise. The network is then tasked with estimating the noise within these corrupted images.

Our model of choice for the denoising network is the U-net. The U-net is one of the most common architectures for denoising \cite{9360532,LEE202092,heinrich2018residual,komatsu2020comparing} and it is often implemented in diffusion models \cite{dhariwal2021diffusion}. Originally introduced for semantic segmentation \cite{U-net}, the U-Net architecture has gained widespread popularity and found applications in diverse image manipulation tasks. The network comprises a downsampling sequence of layers, followed by an upsampling sequence while incorporating skip connections between layers of the same size. Our implementation employs bilinear interpolation for the upsampling process and utilizes average pooling for the downsampling procedure. 

The U-Net configuration can be determined by defining the number of downsampling blocks and the number of channels for each block. The upsampling structure follows a symmetric pattern and the spatial dimension is dependent on the number and resolution of the input and output images. Our U-net utilize 4 downsampling blocks with increasing number of channels, with the number channels being 64, 128, 256, 384. With the number of channels increasing as the spatial dimension decrease. This dimension of U-net proved to be the most effective experimentally, and amounts to around 20 million parameters. To improve the sensibility of the U-net to the noise variance, $\alpha_t$ is taken as input, which is then embedded using an ad-hoc sinusoidal transformation by splitting the value in a set of frequencies, in a way similar to positional encoding in Transformers \cite{attention}. The embedded noise variance is then vectorized and concatenated to the noisy images along the channel axes before being passed to the U-Net. This helps the network to be highly sensitive to the noise level, which is crucial for good performance. 

Conditioning of the model is necessary to guide the diffusion towards a forecast defined by the known previous weather conditions. Our conditioning is applied in a classifier-free manner, by concatenating the conditioning frames to the noisy images alongside the channel axis. Practically, the model $\epsilon_\theta(x_t, t, y)$ takes as input the noisy images $x_t$, which represents the high resolution wind speed at time $t_0$. The conditioning information $y = \{w_{-6}, w_{-3}, w_{0}, w_{+3}\}$ contains the low resolution wind timestamps at -6 hours, -3 hours, current time and +3 hours ahead. 

The input fed into our denoising U-net consists of four preemptively upsampled ERA5 images, along with a single image that represents the noisy input. This configuration outputs a singular, denoised image. Given that the image size is set at 256x256 pixels, the input dimension is structured as (batch\_size, 256, 256, 5), while the output dimension is formulated as (batch\_size, 256, 256). Our implementation directly provides the U-net with the conditioning information, specifically, each temporal slice in the input data is treated analogously to a color channel in an RGB image. 

\subsection{Ensemble Diffusion}
Similarly to \cite{asperti2023precipitation}, in our work we leverage the ensemble properties of diffusion models. With the goals of diffusion models being the approximation of the distribution of the training data, it is expected that our generative outcome is the outcome of a probabilistic point in the data distribution. Thus, it is common that for diffusion processes with the same conditioning the results may vary considerable with different executions. This behaviour of diffusion models may impair the prediction effort, because a single generation may diverge considerably from the mean of the distribution, producing a convincing although highly unlikely outcome inside of the distribution. A possible solution to this problem lies in computing the mean of a set of  diffusion generations, thus moving our generated image close to the mean value of the distribution and therefore generating a more probable outcome. 

In our experiments we settled for an ensemble numbering 15 executions of the single diffusion model, with each diffusion being comprised of 5 steps. This combination proved to be optimal configuration with respect of our testing years. Unlike \cite{asperti2023precipitation}, our approach simply calculates the mean between the predicted instances instead of utilizing a post-processing model; this simpler approach showed comparable results with a lower computational overhead. 

In Figure \ref{fig:big_diffusion} a general overview of the ensemble architecture for evaluation is reported. Starting from the left, we showcase that the inputs are the noise and 4 low-resolution conditioning images. Preemptive bilinear upsampling is performed on the low-resolution images before the insertion into the model. Next, the model performs 5 diffusion steps of sequential denoising, with a noise scheduler managing the re-addition of the noise and the computation of noise rates. This operation is performed 15 times as part of the ensemble effort, using the same conditioning information but different input noise. The cumulated outputs are finally combined via a mean operation to produce the downscaled output. This operation is effectively performed in a parallelized manner with a batch size of 32. 

\subsubsection{Training and Evaluation}

\begin{figure}[h]
\centering
\includegraphics[width=\linewidth]{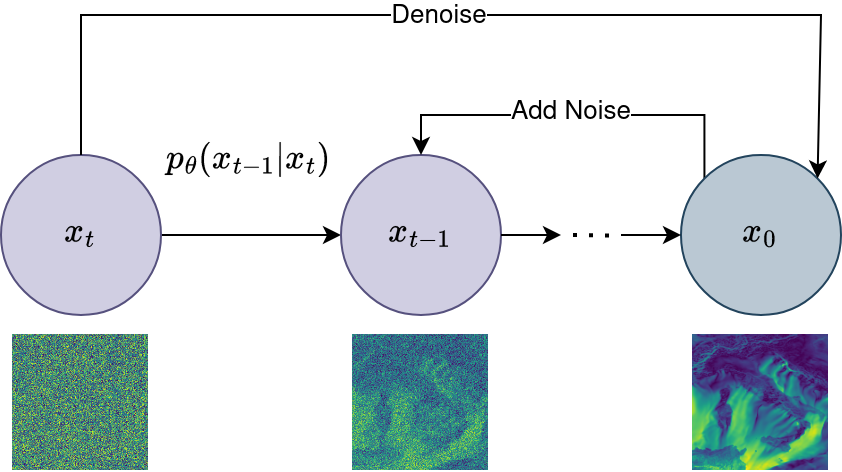}
\caption{The reverse diffusion process is a sequential method that involves alternating between denoising steps and the re-injection of noise, progressively refining the generated image.}
\label{fig:reverse_image}
\end{figure}

The training procedure of diffusion models, as reported in Algorithm \ref{algorithm1}, is implemented by randomly selecting an amount of noise, which is subsequently added on the input images, with the denoising U-net tasked to denoise the image. For the evaluation phase the process is reversed, applying the so called reverse-diffusion, as reported in Algorithm \ref{algorithm2}. The reverse diffusion process involves a sequence of steps, beginning with the introduction of input noise. This procedure alternates between denoising steps and re-injection of noise, as depicted in Figure \ref{fig:reverse_image}. These steps are managed via a noise scheduler, which dictates a sequence of noise rates following a linear progression. For instance, in a 5-step process, the noise rates might be set as 1, 0.8, 0.6, 0.4, and 0.2, respectively. The process starts with a denoising operation at a noise rate of 1, then noise is re-injected at a rate of 0.8, followed by another denoising step at the same rate, and the pattern continues through the sequence.

Our model is developed using the TensorFlow/Keras framework. We trained the model on data from 2010 to 2019, and tested it on data from 2009 and 2020, which allowed us to verify its performance across time periods before and after our training years. The training employs a data generator for efficient handling of training batches, creating custom sequences on the fly, and uses memory mapping to load only the necessary data into RAM. The selection of data batches is randomized over the ten-year training period. The training was conducted in batches of 8 over 200 epochs, using the AdamW optimization algorithm. To optimize training, we implemented a callback function that adjusts both the learning rate and weight decay. The learning rate starts at 1e-04 and gradually reduces to 1e-05, while weight decay starts at 1e-05 and decreases to 1e-06. The training loss, coherently with most diffusion model implementation utilized MAE computed on the noise differences. In the evaluation phase, we employed a generator to produce continuous sequences for each testing year. We configured the batch size to be 32. Taking into account the 3-hour time resolution, this approach resulted in batches representing 96-hour (4-day) periods.

\subsection{Comparative Models}
In our work we selected a set of well known neural models to compare our diffusion architecture with. Our selection include, Efficient Sub-Pixel Convoluted Neural Network (ESPCNN), Enhanced Deep Super-Resolution Network (EDSR) and the residual U-net.

\begin{figure}[h]
\centering
\includegraphics[width=\linewidth]{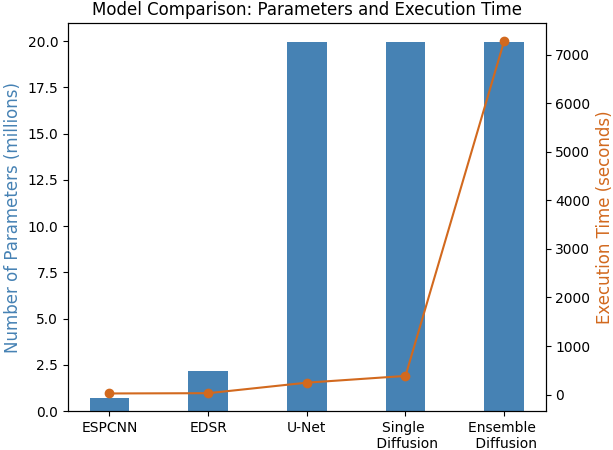}
\caption{Comparative analysis of the tested models highlighting their number of parameters and corresponding execution times for the computation of 1 year of data}
\label{fig:comparative_specs}
\end{figure}

ESPCNN is a well known neural architecture for super-resolution, which utilize sub-pixel convolution as a form of efficient post-upsampling at the very end of the network. This design choice enables the convoluted portion of the network to function at a lower resolution feature dimension, significantly enhancing the network's efficiency. As a result, ESPCNN maintains a low parameter count, making it not only resource-efficient but also faster in processing. This efficient architecture ensures that ESPCNN delivers high-quality super-resolution images while requiring fewer computational resources, a key advantage in practical applications.

The Enhanced Deep Super-Resolution Network (EDSR) is a well known architecture in the realm of super-resolution, renowned for its versatility and effectiveness across a broad spectrum of tasks. EDSR architecture is distinguished by its assembly of multiple residual blocks. A key innovation in EDSR is its use of constant scaling layers, which replace the traditional batch normalization layers. EDSR's design, focused on optimizing image detail and quality, has solidified its status as a top-tier solution in the ever-evolving field of super-resolution technology.

The residual U-net is a U-net implementation which is internally used by the diffusion model to perform the denoising part. The network utilize residual blocks and implements downscaling via average pooling and upsampling via interpolative resizing. It also excels as a standalone super-resolution model. Its effectiveness in this setting is noteworthy, providing an effective benchmark against the diffusion model. This comparison is particularly significant due to the architectural parallels and similarities in parameter count between the two models. Such shared characteristics allow for a more accurate understanding of their respective strengths and capabilities in the field of super-resolution. 

In Figure \ref{fig:comparative_specs} we report a comparison between the tested models, reporting for each one the number of parameters and the execution time necessary to compute one year of downscaled data.

\begin{figure*}[h]
\centering
\includegraphics[width=\linewidth]{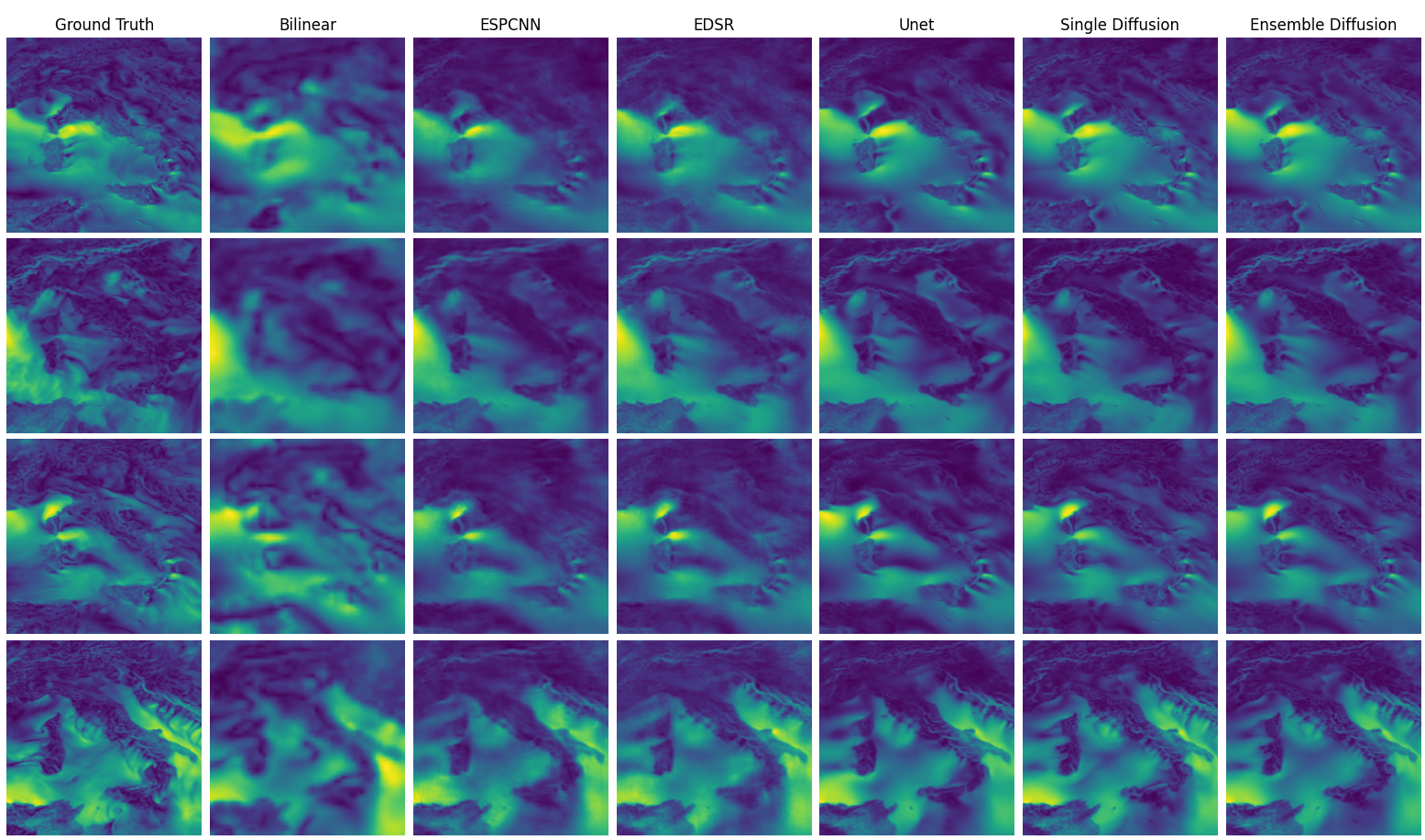}
\caption{General comparison of super resolution methods for 4 randomly selected sequences of the year 2020.}
\label{fig:big_comparison}
\end{figure*}

\subsection{Validation}
The validation process utilizes in-situ observations from weather stations as the ground truth against which the model predictions are evaluated. This process also results in an assessment of the quality of the reanalysis proposed by ERA5 and CERRA.

The IGRA V2 dataset provides for each measurement the value of the wind speed at different pressure levels, where the maximum value coincides with the altitude of the weather station. To remain consistent with the ERA5 and CERRA data, only the observation with the highest air pressure value was selected for each measurement.

Since the data available for the other datasets refer to measurements relative to the previous hour and with 3-hour intervals, the in-situ data considered must also follow a coherent structure. Therefore, given the irregular timestamp distribution of IGRA data, being mainly concentrated around midnight and midday, we decided to remove all detections that did not occur in these two time slots. For what concerns multiple measurements carried out by the same meteorological station during the hour preceding these two slots, these have been reduced to a single average value and the related timestamp rounded to the stroke of the following hour.

To allow a spatially coherent comparison we decided to map all the values to the spatial representation of the in-situ data, in order to keep track of only the values present in ERA5, CERRA and in the results of the models for which there is a correspondence in the validation dataset, thus greatly reducing the space required.

\begin{figure*}[h]
\centering
\includegraphics[width=0.85\linewidth]{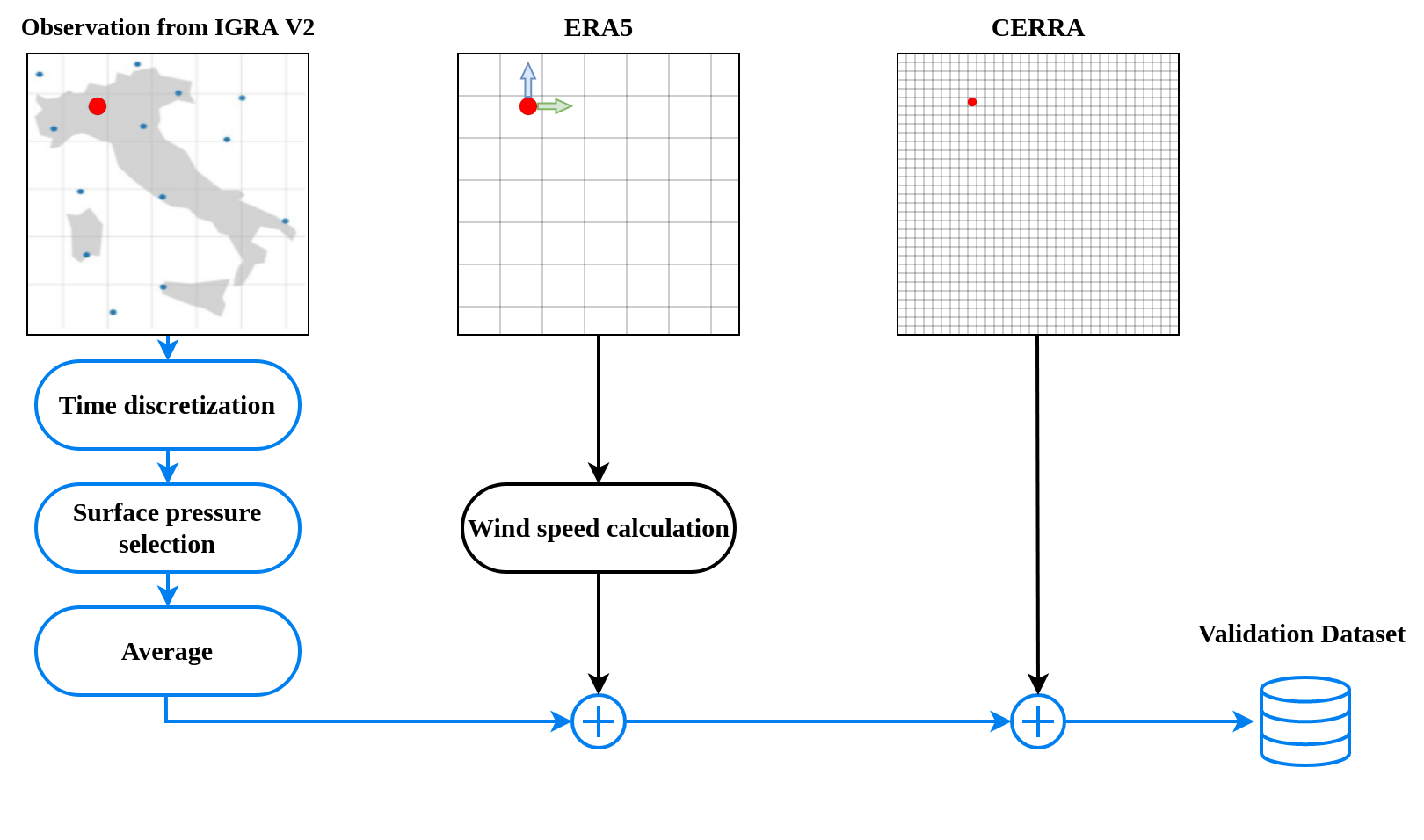}
\caption{Construction process of the dataset used for the validation of ERA5, CERRA and the generated data. IGRA V2 observations in the reference period undergo filtering depending on the reference time and the air pressure value, after which the values selected for an observation at a given time are averaged and become part of the dataset. At this point, the corresponding values of ERA5 (after calculating the wind speed from the two \textit{u} and \textit{v} components) and CERRA are concatenated to each observation. The same process is implemented for the values resulting from the various generative models.}
\label{fig:recap_val}
\end{figure*}

The procedure for assembling the validation dataset is described in Figure \ref{fig:recap_val}.
The obtained dataset can be used to calculate the difference between the results obtained from the various models used to approximate CERRA and the ground truth represented by the observations. Although this does not represent our main objective, this type of analysis also allows us to obtain valuable information on the performance of ERA5 and CERRA themselves as reanalysis datasets.

\section{Experiments and Results}
In this section, we detail the experiments conducted using the models and data referenced in the preceding sections and provide an analysis of the results obtained.

For conducting our experiments, we strategically chose two distinct years, 2009 and 2020, as our testing periods. For both these years, we utilized the lower resolution ERA5 dataset as our conditioning alongside the CERRA dataset, which acted as our high-resolution ground truth. Additionally, we leveraged the IGRA V2 in-situ dataset, containing ground-based observations from meteorological stations, for validating the data generated by our experiments. 

The primary objective of our experiment is to evaluate whether a neural-network-based super-resolution model can effectively replace a traditional downscaling model, like the physics-inspired HARMONIE model currently employed for the generation of CERRA. A key advantage of our neural model is its independence from supplementary data requirements, alongside significantly reduced computational demands. Our comparative analysis studies five competing neural models: ESPCNN, EDSR, U-net, Single Diffusion, and Ensemble Diffusion. This assessment aims to determine not only the feasibility but also the relative performance of these neural models in the context of super-resolution tasks for weather variables.

All the models are trained on a 10 year period with a sequence to one approach, utilizing 4 conditioning low resolution time frames, as described in Section \ref{sec:preprocessing}. The models are then evaluated on the two respective years. A summary of the obtained results is reported in Table \ref{tab:windresults}, where we can note that all neural models have a considerably better performance with respect to our traditional downscaling baseline of bilinear interpolation. The model that achieves the best result, coherently for all three metrics, is Ensemble Diffusion. Ensemble Diffusion in comparison to bilinear interpolation improves the MSE value from 2.50e-03 to 1.02e-03 and provide respectively a 15\% improvement in PSNR and 20\% improvement in SSIM. A set of results for visual inspection is reported in Figure \ref{fig:big_comparison}. 

The data generated through our process underwent validation against in-situ measurements, providing a crucial comparison to determine if the downscaled version offered tangible improvements over real-world observations. As indicated in Table \ref{tab:valresults}, the validation confirmed that our downscaled data align more closely with the actual wind speed measurements in the area, demonstrating enhanced fidelity. 

\begin{table}[htbp]
\centering
\def\arraystretch{1}
\begin{tabular}{>{\bfseries}lccc}
\toprule
\multicolumn{4}{c}{\textbf{Wind speed ERA5 to CERRA Downscaling}} \\
\midrule
\textbf{Model} & \textbf{MSE $\downarrow$} & \textbf{PSNR $\uparrow$} & \textbf{SSIM $\uparrow$}\\ 
\midrule
\multicolumn{4}{c}{2009} \\
\midrule
Bilinear           & 2.50e-03 & 26.36 & 0.708 \\
ESPCNN             & 1.31e-03 & 29.11 & 0.773 \\ 
EDSR               & 1.20e-03 & 29.48 & 0.796 \\ 
U-net              & 1.16e-03 & 29.65 & 0.819 \\ 
Single Diffusion   & 1.18e-03 & 29.62 & 0.829 \\ 
Ensemble Diffusion & \underline{1.07e-03} & \underline{30.10} & \underline{0.844} \\ 
\midrule
\multicolumn{4}{c}{2020} \\
\midrule
Bilinear           & 2.38e-03 & 26.67 & 0.712 \\
ESPCNN             & 1.25e-03 & 29.31 & 0.775 \\ 
EDSR               & 1.14e-03 & 29.70 & 0.797 \\ 
U-net              & 1.10e-03 & 29.86 & 0.820 \\ 
Single Diffusion   & 1.13e-03 & 29.84 & 0.831 \\ 
Ensemble Diffusion & \underline{1.02e-03} & \underline{30.32} & \underline{0.845} \\ 

\bottomrule
\end{tabular}
\caption{Results comparison for multiple models on wind downscaling for the year 2020}
\label{tab:windresults}
\end{table}

\begin{table}[htbp]
\centering
\def\arraystretch{1}
\begin{tabular}{>{\bfseries}lccc}
\toprule
\multicolumn{3}{c}{\textbf{Wind speed validation for the year 2009}} \\
\midrule
\textbf{Model} & \textbf{MAE $\downarrow$} & \textbf{MSE $\downarrow$}\\ 
\midrule
ERA5               & 2.04 & 8.45 \\
CERRA              & \underline{1.86} & \underline{7.39} \\ 
\midrule
Bilinear           & 1.96 & 7.92 \\
ESPCNN             & 1.90 & 7.45 \\ 
EDSR               & 1.92 & 7.63 \\ 
U-net              & 1.88 & 7.42 \\ 
Single Diffusion   & 1.90 & 7.59 \\ 
Ensemble Diffusion & \underline{1.87} & \underline{7.41} \\ 

\bottomrule
\end{tabular}
\caption{Validation of the results obtained compared to the observations present in the IGRA V2 dataset. The comparison takes into consideration the Mean Absolute Error (MAE), measured in meters per second, and the Mean Squared Error (MSE) calculated on the in situ observations and their corresponding values generated by the models or present in ERA5 or CERRA.}
\label{tab:valresults}
\end{table}


For the evaluation dataset, two error measures were taken into consideration to evaluate the closeness of the values produced with respect to the real ones: the Mean Absolute Error (MAE) and the Mean Squared Error (MSE). MAE was chosen for its physical meaning, considering it's measured in meter per second, it can give a clearer understanding of entity of the error, while MSE is more coherent with the results obtained from the neural models. 

Our validation analysis revealed that CERRA outperforms ERA5 in terms of real-world data correlation, achieving a 12.54\% improvement in Mean Squared Error (MSE). Among the downscaling models we tested, the Ensemble Diffusion method emerged as the most effective, showing a 12.30\% performance enhancement over ERA5 and closely trailing CERRA by a mere 0.27\% in terms of MSE. These findings underscore the value added by the downscaled data and highlight the Ensemble Diffusion model's capability to nearly match the performance of CERRA. These results showcase the capabilities of neural models to effectively approximate the inner mechanics of state of the art physics inspired model, providing both lower computational costs and avoiding the need for additional information that are often difficult do obtain, with common delays in the publication of current data. Our validation was possible only for the year 2009, as validation data was not available for the year 2020. 


\begin{figure*}[h]
\centering
\includegraphics[width=0.9\linewidth]{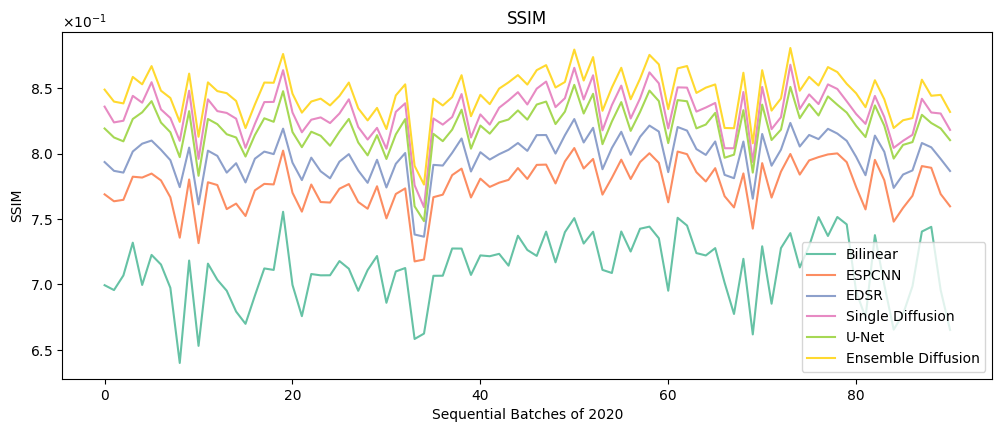}
\caption{Graph comparison for the SSIM metric of the different tested models. SSIM is calculated for each batch sequentially in the testing year of 2020, highlighting the variabilty of performance at different times of the year. }
\label{fig:ssim_graph}
\end{figure*}

The Structural Similarity Index (SSIM) is widely acknowledged as a crucial metric for evaluating super-resolution performance, and we are interested to explore the changes in its performance within the training year. Figure \ref{fig:ssim_graph} presents a comparative analysis of SSIM metrics across the various downscaling methods for the year 2020. Notably, there is a significant fluctuation in performance throughout the year. These variations, consistently observed across all models including bilinear interpolation, suggest that temporal weather variations significantly influence the efficacy of the downscaling models. This consistent trend across different models implies that the performance changes are predominantly driven by weather-related factors rather than the intrinsic characteristics of the models themselves.

\begin{figure}[h]
\centering
\includegraphics[width=\linewidth]{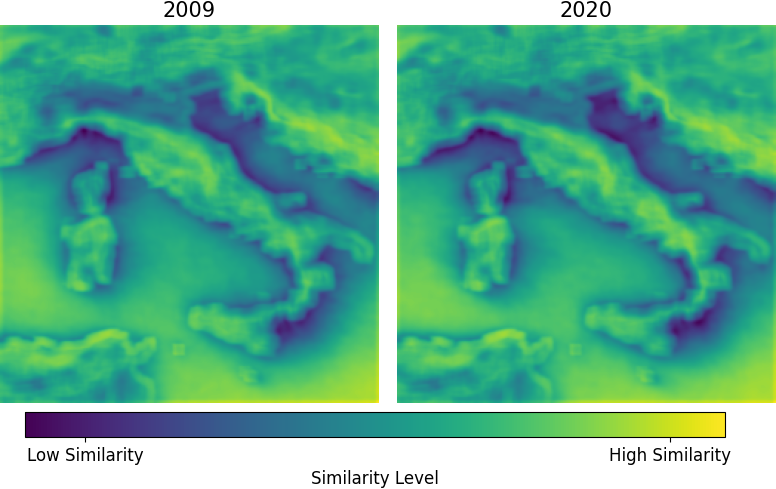}
\caption{Mean image representing the spatial SSIM error for both testing years of 2009 and 2020, showcasing the downscaling performance relative to different geographical areas.}
\label{fig:ssim_mean}
\end{figure}

Furthermore, an in-depth geographical analysis of SSIM performance can provide insights into which areas are harder to effectively downscale. Given that SSIM is calculated over patches, it's feasible to compile a comprehensive SSIM output. This output takes the form of a matrix, where each pixel value denotes the SSIM computed over a window centered on that corresponding pixel in the input images. In Figure \ref{fig:ssim_mean}, we showcase mean images derived from averaging these full SSIM matrices for the entire testing years of 2009 and 2020, respectively. These images reveal that coastal regions typically exhibit higher error rates, with the most pronounced discrepancies observed in the high Adriatic Sea, the Strait of Messina, and the Ligurian Sea. The results also depict that the geographical errors are coherent between the two testing years. 

As a final experiment, considering the unavaliability of CERRA for the years ranging from 2021 to 2024, we also make publicly available for further studies our computed approximation of CERRA, computed via Ensemble Diffusion.


\section{Discussion \& Conclusion}
In this research we successfully employed state of the art super-resolution diffusion models to downscale wind speed data from ERA5 to CERRA. This approach approximates the existing model used for generating CERRA, but it innovatively does so in a data-driven manner without the need for supplementary data. Our experiment was focused on the Mediterranean area surrounding Italy and utilizing wind speed as our main weather variable. 

Our experiments, conducted over two distinct testing years, demonstrated that the downscaling computed by our method aligns closely with the outcomes of traditional physics-based models. This alignment was evident both in the similarity of results and in their concordance with measurements from ground-based in-situ stations. Among all the models tested, the Ensemble Diffusion approach achieved the most accurate results. However, it's noteworthy that even less complex super-resolution models consistently outperformed basic methods like bilinear interpolation. Our findings reveal the practicality of using super-resolution diffusion  models in providing timely and informative downscaled data for meteorological studies. Given the significant delays often encountered in downscaled reanalyses – for example, CERRA's data lagging over two years behind the current date at the time of this study – our findings are particularly relevant for the scientific community. These delays typically arise from the unavailability of additional information needed by physics-based models, which depend on various sources and are subject to time-consuming computational processes. Our approach, leveraging a neural model trained on previously generated data, offers a timely alternative for accessing current downscaled results. This method not only reduces computational costs but also maintains a high quality of output. Importantly, our model produces data that aligns more closely with actual measurements than the original low-resolution data, demonstrating its practical potential in enhancing real-time weather analysis.

In our study, we encountered differing time scales between the high-resolution dataset (CERRA) and the low-resolution dataset (ERA5). ERA5 offers hourly temporal resolution, whereas CERRA provides data at a 3-hour resolution. Given that our model can generate sequences conditioned only on ERA5 data, we can utilize our model to augment the temporal resolution of our downscaled data to match the temporal resolution of ERA, providing a higher temporal resolution when compared to the physics-inspired reanalyses model. 

Regarding the main limitations of our approach, it's important to note that our data-driven method requires pre-existing downscaled data for training, thus relying on already established models. Additionally, our experiments represent a preliminary phase, focusing solely on a single weather variable and a relatively limited geographical area. Future developments could encompass a broader range of weather variables and potentially cover the entire region addressed by CERRA. This expanded scope could lead to improved results, as the model would gain a more comprehensive understanding of the weather system with access to diverse meteorological data. Moreover, hardware constraints limited our training dataset to only 10 years, which is noteworthy given the high quality of our results. In future research, it would be beneficial to utilize a more extensive dataset, potentially spanning the full 40 years available from CERRA, to enhance the model's accuracy and predictive capabilities.

In conclusion, our research presents a viable alternative for obtaining downscaled data of current events. We demonstrate the potential of diffusion models to effectively approximate the processes of traditional downscaling models, avoiding delays in data availability. Furthermore, our findings suggest a promising future for reanalysis models, potentially shifting towards neural network-based approaches that leverage the advantages identified in this study. This shift could mark a significant advancement in the field, offering more efficient and accurate ways to handle meteorological data.

\section{Acknowledgements}
ERA5 and CERRA were downloaded from the Copernicus Climate Change Service (C3S) (2023).
This research was partially funded and supported by the following Projects:
\begin{itemize}
\item European Cordis Project ``Optimal High Resolution Earth System Models for Exploring Future Climate Changes'' (OptimESM),
Grant agreement ID: 101081193
\item Future AI Research (FAIR) project of the National Recovery and Resilience Plan (NRRP), Mission 4 Component 2 Investment 1.3 funded from the European Union - NextGenerationEU.
\item ISCRA Project ``AI for weather analysis and forecast'' (AIWAF)
\end{itemize}

\subsection*{Statements and Declarations}
The authors declare no competing interests.

\subsection*{Code availability}
The code and downscaled data relative to the presented work is archived at the following GitHub \href{repository}{https://github.com/fmerizzi/ERA5-to-CERRA-via-Diffusion-Models/}.





\bibliography{bibliography,forecasting,superres}

\end{document}